\documentclass[letterpaper, 10 pt, conference]{ieeeconf}  

\IEEEoverridecommandlockouts                              

\usepackage{xspace}

\usepackage{times}
\usepackage{epsfig}
\usepackage{graphicx}
\usepackage{amsmath}
\usepackage{amssymb}
\usepackage[outercaption]{sidecap}     

\usepackage[colorlinks=true,allcolors=green]{hyperref}

\usepackage{etoolbox}
\makeatletter
\patchcmd{\@makecaption}
  {\scshape}
  {}
  {}
  {}
\makeatother

\title{\LARGE \bf
Real-time 3D Traffic Cone Detection for Autonomous Driving
}

\author{Ankit Dhall, Dengxin Dai and Luc Van Gool%
\thanks{All authors are with ETH-Zurich
        {\tt\small adhall@ethz.ch, \{dai, vangool\}@vision.ee.ethz.ch}}%
}

\begin{document}

\maketitle
\thispagestyle{empty}
\pagestyle{empty}

\begin{abstract}
Considerable progress has been made in semantic scene understanding of road scenes with monocular cameras. It is, however, mainly focused on certain specific classes such as cars, bicyclists and pedestrians. This work investigates traffic cones, an object category crucial for traffic control in the context of autonomous vehicles. 
3D object detection using images from a monocular camera is intrinsically an ill-posed problem. In this work, we
exploit the unique structure of traffic cones and propose a pipelined approach to solve this problem. Specifically, we first detect cones in images by a modified 2D object detector. Following which the keypoints on a traffic cone are recognized with the help of our deep structural regression network, here, the fact that the cross-ratio is projection invariant is leveraged for network regularization. Finally, the 3D position of cones is recovered via the classical Perspective n-Point algorithm using correspondences obtained form the keypoint regression.   
Extensive experiments show that our approach can accurately detect traffic cones and estimate their position in the 3D world in real time. The proposed method is also deployed on a real-time, autonomous system. It runs efficiently on the low-power Jetson TX2, providing accurate 3D position estimates, allowing a race-car to map and drive autonomously on an unseen track indicated by traffic cones. With the help of robust and accurate perception, our race-car won both Formula Student Competitions held in Italy and Germany in 2018, cruising at a top-speed of 54 km/h on our driverless platform ``gotthard driverless'' ~\url{https://youtu.be/HegmIXASKow?t=11694}. Visualization of the complete pipeline, mapping and navigation can be found on our project page ~\url{http://people.ee.ethz.ch/~tracezuerich/TrafficCone/}.
\end{abstract}

\section{Introduction}
Autonomous driving has become one of the most interesting problems to be tackled jointly by the computer vision, robotics and machine learning community~\cite{AD:Boss:08,AD:systems:algorithm:11,drive:surroundview:route:planner,human-like:driving:19}. Numerous studies have been done to take the field of autonomous driving forward, leading to ambitious announcements promising fully automated cars in a couple of years. Yet, significant technical challenges
such as the need for necessary robustness against adverse weather and changing illumination conditions~\cite{semantic:foggy:scene,daytime:2:nighttime,valada2017adapnet}, or the capability to cope with temporary, unforeseen situations such as roadside construction and accidents~\cite{problems:autonomous:vehicle:17} must be overcome before a human driver can make way for autonomous driving. 

Traffic control devices, such as traffic signs, traffic lights and traffic cones, play a crucial role in ensuring safe driving and preventing mishaps on the road. Recent years have witnessed great progress in detecting traffic signs~\cite{Ruta2011,traffic:sign:solution:13} and traffic lights~\cite{traffic:light,three:ways:stereo:vision:tl}. Traffic cones, however, have not received due attention yet. Traffic cones are conical markers and are usually placed on roads or footpaths and maybe used to safely and temporarily redirect traffic or cordon-off an area. They may often be used for splitting or merging traffic lanes in the case of roadside construction and automobile accidents. These situations need to be addressed internally with on-board sensors because even high-definition (HD) maps cannot solve this problem as the traffic cones are temporary and movable.

\begin{figure}[!bt]
    \centering
    \includegraphics[width=0.47\textwidth]{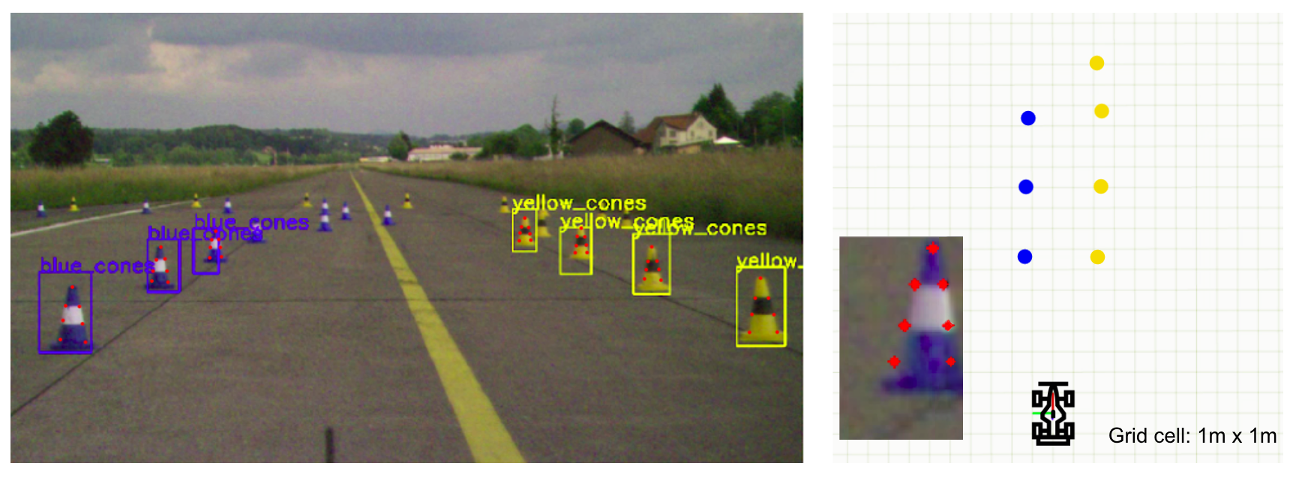}
    \caption{The pipeline can be subdivided into three parts: (1) object detection, (2) keypoint regression and (3) 2D-3D correspondence followed by 3D pose estimation from a monocular image.}
    \label{fig:pipeline_overview}
\end{figure}

It may be tempting to employ an end-to-end approach to map directly from images to control outputs (throttle and steering commands) \cite{drive:surroundview:route:planner}. We, however, believe that a fusion of part-based approaches with interpretable sub-modules and the data-driven end-to-end methods is a more promising direction. Object detection in any case is still very necessary for learning autonomous driving systems.


It is interesting to note here that although these traffic cones are static objects themselves, they are frequently replaced and moved around the urban driving scenario. Cars may break down unexpectedly and new constructions zones may pop up more often than anticipated. Although, buildings and landmarks can be mapped with ease and used for localization, one needs to actively detect and estimate the position of these traffic cones for safe, automated driving.



A range based sensor, such as the LiDAR is designed to accurately measure 3D position, but because a LiDAR has a sparse representation as compared to an image detecting small objects and predicting about their physical properties such as their color and texture becomes a massive challenge. Additionally, LiDAR sensors are more expensive than cameras, driving the costs of such platform to the higher end of the spectrum. Advances in computer vision show that images from even a monocular camera can be used to not only reveal \textit{what} is in the scene but also \textit{where} it is physically in the 3D world \cite{glasner2012aware, vp_and_kp}. Another advantage of using a monocular camera is that a multi-camera setup is not required, making the system more cost-effective and maintainable.

In this work, we tackle 3D position estimation and detection of traffic cones from a single image. 
We break the task into three steps: 2D object detection, regressing landmark keypoints, and finally mapping from the 2D image space to 3D world coordinates. 
In particular, cones are detected in images by an off-the-shelf 2D object detector customized for this scenario; the detected 2D bounding boxes are fed into our proposed convolutional neural network to regress seven landmark keypoints on the traffic cones, where the fact that cross-ratio ($Cr$) is projection invariant is leveraged for robustness. Finally, the 3D position of cones is recovered by the Perspective n-Point algorithm. In order to train and evaluate our algorithm, we construct a dataset of our own for traffic cones.

Through extensive experiments we show that traffic cones can be detected accurately using single images by our method. The 3D cones position estimates deviate by only 0.5m at 10m and 1m at 16m distances when compared with the ground-truth. We further validate the performance of our method by deploying it on a critical, real-time system in the form of a life-sized autonomous race-car. The car can drive at a top-speed of 54 km/h on a track flanked by traffic cones. 


The main contribution of this paper are (1) a novel method for real-time 3D traffic cone detection using a single image and (2) a system showing that the accuracy of our pipeline is sufficient to autonomously navigate a race-car at a top-speed of 54 km/h. 
The video showing our vehicle navigating through a track flanked by traffic cones can be found at \url{https://youtu.be/HegmIXASKow?t=11694}.



\section{Related Work}


\textbf{Fast object detection.} Object detection has been one of the most prized problems in the field of computer vision. Moreover, for real-time, on-line performance especially on robotics platforms speed is of the essence. One of the first successful fast object detector is Viola-Jones' face detector \cite{viola2001rapid}, which employs weak learners to accurately detect faces using Haar-based features. 
The next class of well-known object detectors uses deep learning in the form of convolutional neural networks (CNNs). The string of R-CNN \cite{girshick2014rich, ren2015faster, girshick2015fast} schemes use CNN-based features for region proposal classification. YOLO \cite{redmon2016you} cleverly formulates object detection as a regression task, leading to very efficient detection systems. Single shot has been employed to 3D object detection as well~\cite{Kehl_2017_ICCV}. While progress has been made in terms of general object detection, the performance on small-object classes such as traffic cones requires further improvements.   


\textbf{Traffic device detection.} 
Work has been done in the direction of detecting traffic sign~\cite{Ruta2011,traffic:sign:solution:13} and traffic light~\cite{traffic:light,three:ways:stereo:vision:tl}. To aid in the efforts for bench-marking, a 100,000 annotated image dataset for traffic signs has been released \cite{zhu2016traffic}. Li et al. \cite{li2017perceptual} propose a generative adversarial network (GAN) to improve tiny object detection, such as distant traffic signs.
Lee et al. \cite{lee2018simultaneous} explore the idea of detecting traffic signs and output a finer mask instead of a coarse rectangle in the form of a bounding box. The work briefly discusses triangulation of points using the extracted object boundary across 2 frames, but is limited as it is only in simulation. Our work focuses on traffic cone detection and 3D position estimation using only a single image. 

\textbf{Keypoint estimation.} One of the main contributions of this work is to be able to accurately estimate the 3D pose of traffic cones using just a single frame. 
A priori information about the 3D geometry is used to regress highly specific feature points called \textit{keypoints}. Previously, pose estimation and keypoints have appeared in other works \cite{savarese20073d, glasner2012aware}. Glasner et al. \cite{glasner2012aware} estimate pose for images containing cars using an ensemble of voting SVMs. Tulsiani et al. \cite{vp_and_kp} use features and convolutional neural networks to predict view-points of different objects. Their work captures the interplay between viewpoints of objects and keypoints for specific objects. PoseCNN \cite{posecnn} directly outputs the 6 degrees-of-freedom pose using deep learning. Gkioxari et al. \cite{gkioxari2014using} use a k-part deformable parts model and present a unified approach for detection and keypoint extraction on people. Our method leverages the unique structure of traffic cones, more specifically the projective invariant property of cross-ratio, for robust keypoint estimation. 



\section{Monocular Camera Perception Pipeline}

\subsection{Sensor Setup and Computation Platform}




\begin{figure}[!tb]
    \centering
    \includegraphics[ height=0.25\textwidth]{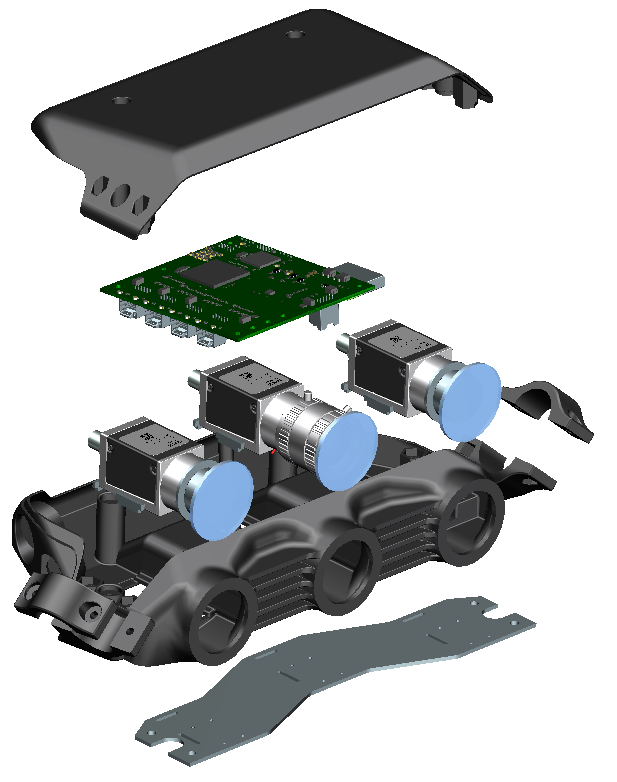}
    \caption{The left and right cameras (on the extremes) in the housing act in a stereo configuration; the center camera is a stand-alone monocular camera and uses the pipeline elaborated in this work.}
    \label{fig:housing}
\end{figure}

The experimental setup consists of 2-megapixel cameras (CMOS sensor-based) with a global shutter to avoid image distortion and artifacts due to fast motion.  Figure \ref{fig:housing} shows our camera setup. 
The center camera, which is the monocular camera described in this work, has a 12mm lens to allow long range perception. The left and right cameras use lenses with a 5.5mm focal length and act as stereo pair for triangulating cones close-by.
The cameras are enclosed in a customized 3D printed, water-proof shell with polarized filters in front of the lenses. The cameras transmit data over ethernet through a shared switch, allowing for a neat camera housing. The cameras are screwed to the metallic plate at the bottom and are in direct contact to keep them at operating temperatures. Raw camera data is directly transmitted to a Jetson TX2 which acts as a slave to the master PIP-39 (with Intel i7) onboard ``gotthard driverless''. The pipeline is light enough to run completely on a low-powered computing at a rate of 10Hz.

\subsection{Pipeline Overview}
Pose estimation from a single image is an ill-posed problem but it is solvable with a priori structural information of the object of interest.
With the availability of tremendous amounts of data and powerful hardware such as GPUs, deep learning has proven to be good at tasks that would be difficult to solve with classical, hand-crafted approaches. Data-driven machine learning does well to learn sophisticated representations while results established from mathematics and geometry provide robust and reliable pose estimates. In our work we strive to combine the best of both worlds in an efficient way holding both performance and interpretability in high regard with a pipelined approach.

The sub-modules in the pipeline enable it to detect objects of interest and accurately estimate their 3D position by making use of a single image. The 3 sub-modules of the pipeline are (1) object detection, (2) keypoint regression and (3) pose estimation by 2D-3D correspondence. The pipeline's sub-modules are run as nodes using Robot Operating System (ROS) \cite{ros} framework that handles communication and transmission of data (in the form of messages) between different parts of the pipeline and also across different systems. The details will be described in more detail in Section~\ref{sec:approach}.  

\begin{figure*}[!tb]
    \centering
    \includegraphics[width=.8\textwidth]{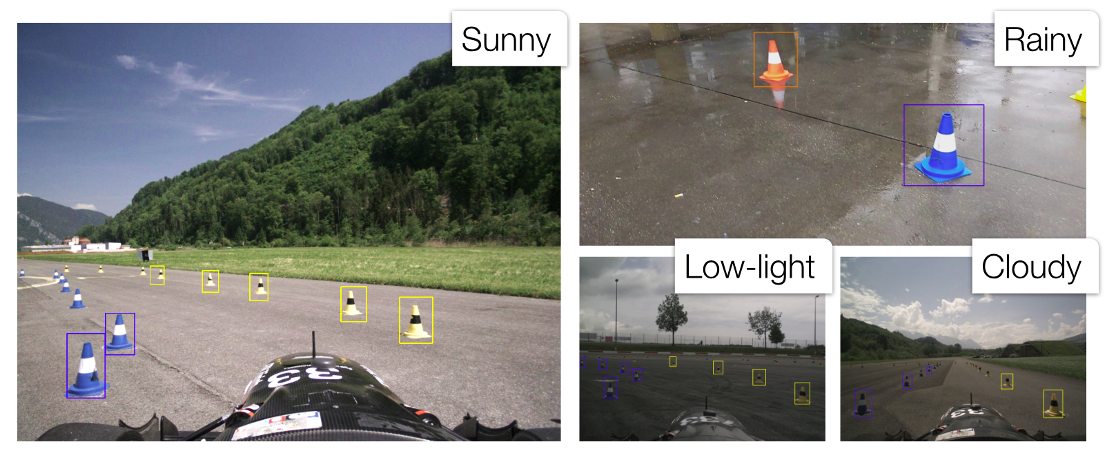}
    \caption{Detection under varying lighting and weather conditions for \textit{yellow}, \textit{blue} and \textit{orange} cones.}
    \label{fig:robust_detection}
\end{figure*}

\section{Approach}
\label{sec:approach} 

\subsection{Object Detection}
\label{sec:object:detection}
To estimate 3D position of multiple object instances from a single image, it is necessary to first be able to detect these objects of interest. For the task of object detection, we employ an off-the-shelf object detector in our pipeline in the form of YOLOv2 \cite{YOLOv2}.
YOLOv2 is trained for the purpose of detecting differently colored cones that serve as principal landmarks to demarcate the race-track at Formula Student Driverless events (where we participated with our platform). Thresholds and parameters are chosen such that false positives and misclassification are minimal. For this particular use-case, YOLOv2 is customized by reducing the number of classes that it detects, as it only needs to distinguish among \textit{yellow}, \textit{blue} and \textit{orange} cones each with a particular semantic meaning on the race-track.

Since the bounding boxes for cones have a height to width ratio of greater than one, such prior information is exploited by re-calculating the \textit{anchor boxes} used by YOLOv2 (see \cite{YOLOv2} for details).



Weights trained for the ImageNet \cite{imagenet} challenge are used for initialization. We follow a similar training scheme as in the original work \cite{YOLOv2}. The detector is fine-tuned when more data is acquired and labeled during the course of the season. Refer to Section \ref{sssec:YOLO_data} for more details on data collection and annotation.

\subsection{Keypoint Regression}
This section discusses how object detection in a single 2D image can be used to estimate 3D positions of objects of interest. Doing this from a single view of the scene is challenging because of ambiguities due to scale. However, with prior information about the 3D shape, size and geometry of the cone, one can recover the 3D pose of detected objects using only a single image. 
One would be able to estimate an object's 3D pose, if there is a set of 2D-3D correspondences between the object (in 3D) and the image (in 2D), along with intrinsic camera parameters.

To this end, we introduce a feature extraction scheme that is inspired by classical computer vision but has a flavor of learning from data using machine learning.

\begin{figure}[!tb]
    \centering
    \includegraphics[width=0.35\textwidth]{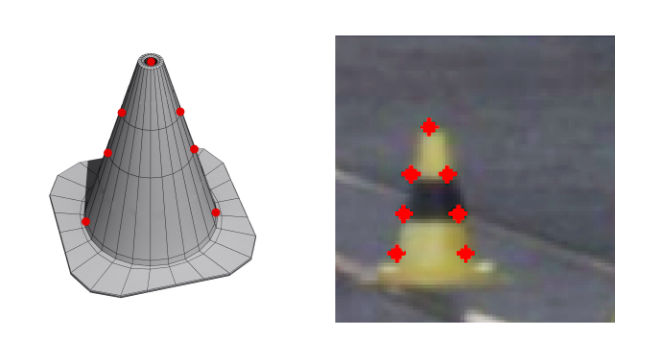}
    \caption{3D model of the cone and a representative sub-image patch with the image of the cone. The red markers correspond to the 7 specific keypoints the keypoint network regresses from a given cone patch.}
    \label{fig:model_and_kp}
\end{figure}

\subsubsection{Keypoint Representation}

The bounding boxes from the object detector do not directly correspond to a cone. To tackle this, we extract landmark features within the proposed bounding box that are representative of the cone. In the context of classical computer vision, there are three kinds of features: \textit{flat regions}, \textit{edges} and \textit{corners}. \textit{Edges} are more interesting than \textit{flat} regions as they have a gradient in one direction (perpendicular to the edge boundary) but suffer from the aperture problem \cite{rich_s}. By far, the most interesting features are the \textit{corners} that have gradients in two directions instead of one making them most distinctive among the three.

Previous feature extraction works include the renowned Harris corners detector \cite{harris1988combined}, robust SIFT \cite{lowe2004distinctive} and SURF \cite{bay2006surf} feature extractors and descriptors.
A property that many of these possess is invariance to transformations such as scale, rotation and illumination, which for most use-cases is quite desirable. Most of these work well as general feature detectors and can be used across a range of different applications.

The issue with using such pre-existing feature extraction techniques is that they are generic and detect any kind of features that fall within their criteria of what a feature point is. For instance, a Harris corner does not distinguish whether the feature point lies on a cone or on a crevasse on the road. This makes it difficult to draw the relevant 2D correspondences and match them correctly to their 3D counterparts. Another issue is when a patch has a low resolution, it may detect only a couple of features which will not provide enough information to estimate the 3D pose of an object.

\subsubsection{Keypoint Regression}


With these limitations of previously proposed work in mind, we design a convolutional neural network (CNN) that regresses ``corner'' like features given an image patch. The primary advantage over generic feature extraction techniques is that with the help of data one can learn to robustly detect extremely specific feature points. Although, in this work we focus on a particular class of objects, the cones; the proposed keypoint regression scheme can be easily extended to different types of objects as well. 
The 3D locations corresponding to these specific feature points (as shown in Figure \ref{fig:model_and_kp}) can be measured in 3D from an arbitrary world frame, $\mathcal{F}_w$. For our purpose, we place this frame at the base of the cone.

There are two reasons to have these keypoints located where they are. First, the keypoint network regresses positions of 7 very specific features that are visually distinct and can be considered visually similar to ``corner'' features. Second, and more importantly, these 7 points are relatively easy to measure in 3D from a fixed world frame $\mathcal{F}_w$. For convenience $\mathcal{F}_w$ is chosen to be the base of the 3D cone, enabling measurement of 3D position of these 7 points in this world frame, $\mathcal{F}_w$. The 7 keypoints are the apex of the cone, two points (one on either side) at the base of the cone, 4 points where the center stripe, the background and the upper/lower stripes meet.

The customized CNN, made to detect specific ``corner'' features, takes as input an $80 \times 80 \times 3$ sub-image patch which contains a cone, as detected by the object detector, and maps it to $\mathbb{R}^{14}$. The spatial dimensions are chosen as $80 \times 80$, the average size of detected bounding boxes. The output vector of $\mathbb{R}^{14}$ are the $(x, y)$ coordinates of the keypoints.

\begin{figure}[!tb]
    \centering
    \includegraphics[width=0.47\textwidth]{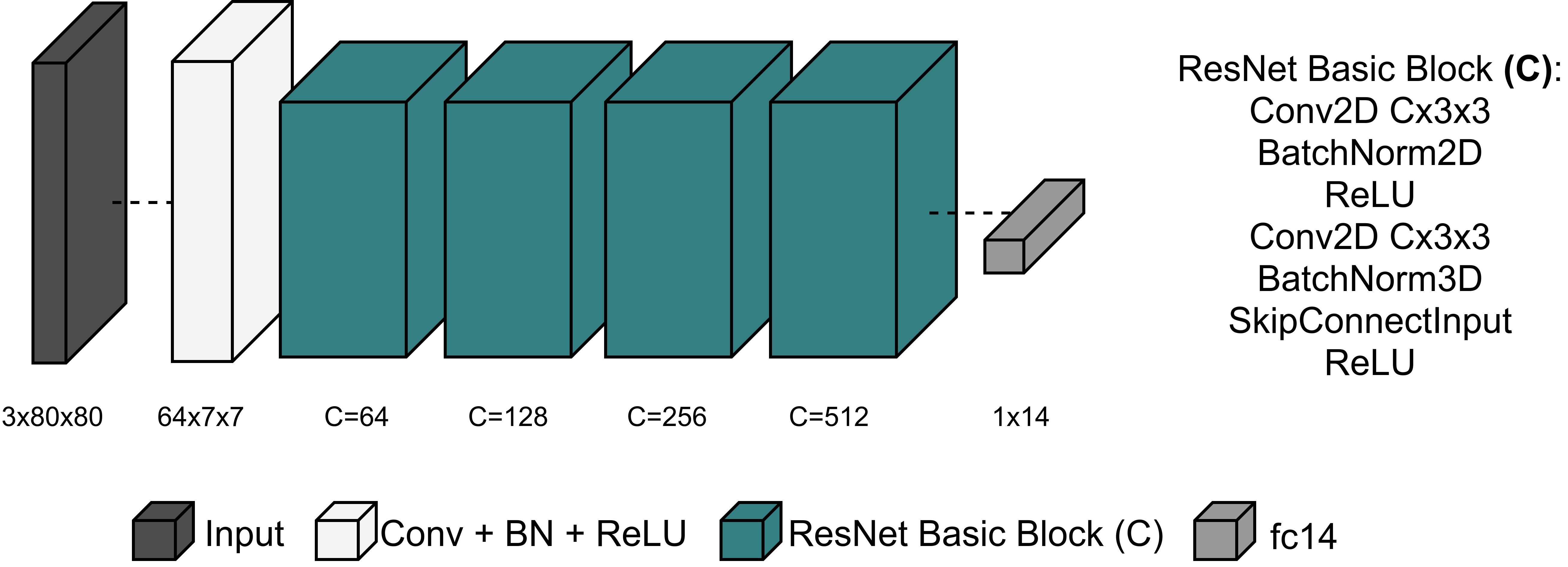}
    \caption{Architecture of the keypoint network. It takes a sub-image patch of $80 \times 80 \times 3$ as input and maps it to $\mathbb{R}^{14}$, the $(x, y)$ coordinates for the 7 keypoints. It can process 45-50 cone patches per second on a low-powered Jetson TX2.}
    \label{fig:network_arch}
\end{figure}

The architecture of the convolutional neural network consists of basic residual blocks inspired by ResNet \cite{he2016deep} and is implemented using the PyTorch \cite{pytorch} framework.

As analyzed in \cite{unet}, with more convolutional layers, the tensor volume has more channels but on the other hand there is a significant reduction in the spatial dimensions, implying the tensors contain more global and high-level information than specific, local information. We eventually care about location of keypoints which are extremely specific and local. Using such an architecture prevents loss of spatial information as it is crucial to predict the position of keypoints accurately.


The backbone of the network is similar to the ResNet. The first block in the network is a convolution layer with a batch norm (BN) followed by rectified linear units (ReLU) as the non-linear activation. The next 4 blocks are basic residual blocks with increasing channels $C \in \{64, 128, 256, 512\}$ as depicted in Figure \ref{fig:network_arch}. Finally, there is a fully-connected layer that regresses the $(x, y)$ position of the keypoints in the patch.


\subsubsection{Loss Function}
\label{sec:loss:function} 
As mentioned previously, we use object-specific prior information to match 2D-3D correspondences, from the keypoints on the image (2D) to their location on the physical cone (3D). In addition, the keypoint network also exploits a priori information about the object's 3D geometry and appearance through the loss function via the concept of the \textit{cross-ratio}.
As known, under projective transform, neither distances between points nor their ratio is preserved. However, a more complicated entity known as the \textit{cross-ratio}, which is the ratio of ratio of distances, is invariant and is preserved under a projection. While used in classical computer vision approaches that involve geometry, cross-ratio has seldom been used in the context of machine learning. We use it to geometrically constrain the location of the keypoints and directly integrate into the model's loss function.

The cross-ratio ($Cr$) is a scalar quantity and can be calculated with 4 collinear points or, 5 or more non-collinear points \cite{crossratio}. It is invariant under a projection and the process of acquiring images with a camera is essentially a projective transform. The cross-ratio is preserved, both on the 2D projection of the scene (the image) and in 3D space where the object lies.

In our case, we use 4 collinear points $p_1, p_2, p_3, p_4$ to calculate the cross-ratio as defined in Equation \ref{eq:cross_ratio}. Depending on whether the value is calculated for 3D points ($D=3$) or their projected two dimensional counterparts ($D=2$), the distance $\Delta_{ij}$, between two points, $p_i$ and $p_j$ is defined.


\begin{align}
\begin{aligned}
Cr(p_1, p_2, p_3, p_4) = (\Delta_{13}/\Delta_{14})/(\Delta_{23}/\Delta_{24}) \in \mathbb{R}\\
\Delta_{ij} = \sqrt{\Sigma_{n=1}^{D} (X_{i}^{(n)} - X_{j}^{(n)})^2}, D \in \{2, 3\}
\end{aligned}
\label{eq:cross_ratio}
\end{align}



In addition to the cross-ratio to act as a regularizer, the loss has a squared error term for the $(x, y)$ location of each regressed keypoint. The squared error term forces the regressed output to be as close as possible to the ground-truth annotation of the keypoints. The effect of the cross-ratio is controlled by the factor $\gamma$ and is set to a value of 0.0001.


\begin{align}
\begin{aligned}
\Sigma_{i=1}^{7} (p_i^{(x)} - p_{i\_groundtruth}^{(x)})^2 + (p_i^{(y)} - p_{i\_groundtruth}^{(y)})^2 \\
+ \gamma \cdot (Cr(p_1, p_2, p_3, p_4) - Cr_{3D})^2 \\
+ \gamma \cdot (Cr(p_1, p_5, p_6, p_7) - Cr_{3D})^2
\end{aligned}
\label{eq:loss_fn}
\end{align}

\begin{figure}[!tb]
    \centering
    \includegraphics[width=0.2\textwidth]{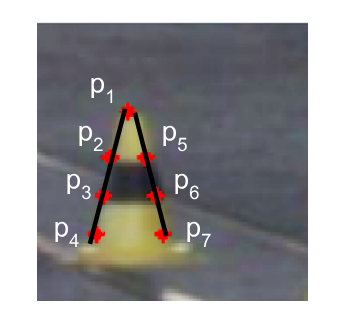}
    \caption{An exemplary $80 \times 80$ cone patch with regressed keypoints overlaid in red. Depiction of the left ($p_1, p_2, p_3, p_4$) and right arm ($p_1, p_5, p_6, p_7$) of the cone. Both of which are used to calculate the cross-ratio terms and minimize the error between themselves and the cross-ratio on the 3D object ($Cr_{3D}$).}
    \label{fig:cross_ratio_on_cone}
    
\end{figure}

The second and third term minimize the error between the cross-ratio measured in 3D ($Cr_{3D}$) and the cross-ratio calculated in 2D based on the keypoint regressor's output, indirectly having an influence on the locations output by the CNN. The second term in Equation \ref{eq:loss_fn} represents the left arm of the cone while the third term is for the right arm, as illustrated in Figure \ref{fig:cross_ratio_on_cone}. For the cross-ratio, we choose to minimize the squared error term between the already known 3D estimate ($Cr_{3D}=1.39408$ from a real cone) and its 2D counterpart. Equation \ref{eq:loss_fn} represents the loss function minimized while training the keypoint regressor. The training scheme is explained in the  following section.

\subsubsection{Training}
To train the model, Stochastic Gradient Descent (SGD) is used for optimization, with $learning\ rate=0.0001$, $momentum=0.9$ and a batch size of 128. The learning rate is scaled by 0.1 after 75 and 100 epochs. The network is trained for 250 epochs. The keypoint regressor is implemented in PyTorch and used via ROS on ``gotthard driverless''. Refer to Section \ref{sssec:kp_data} for more information about the dataset.


\subsection{2D to 3D Correspondences}
The keypoint network provides the location of specific features on the object of interest, the keypoints. 
We use a priori information about the object of interest (the cone, in this case) such as its shape, size, appearance and 3D geometry to perform 2D-3D correspondence matching. The camera's intrinsic parameters are available and the keypoint network provides the 2D-3D correspondences. Using these pieces of information it is possible to estimate the 3D pose of the object in question solely with a single image. We stitch these pieces together using the Perspective n-Point (PnP) algorithm.

We define the camera frame as $\mathcal{F}_c$ and the world frame as $\mathcal{F}_w$. Although $\mathcal{F}_w$ can be chosen arbitrarily, in our case, we choose $\mathcal{F}_w$ to be at the base of every detected cone, for the ease of measurement of the 3D location of the keypoints (with respect to $\mathcal{F}_w$) and convenience of calculating the transform and eventually the cone position.

We use Perspective n-Point to estimate the pose of every detected cone. This works by estimating the transform $^{c}\mathcal{T}_w$ between the camera coordinate frame, $\mathcal{F}_c$, and the world coordinate frame, $\mathcal{F}_w$. As we are concerned only with the translation between $\mathcal{F}_c$ and $\mathcal{F}_w$, which is exactly the position of the cone with respect to the camera frame that we wish to estimate, in our case we discard the orientation.

To estimate the position of the cone accurately, we use non-linear PnP implemented in OpenCV \cite{opencv3Drecon} which uses Levenberg-Marquardt to obtain the transformation. In addition, RANSAC PnP is used instead of vanilla PnP, to tackle and deal with noisy correspondences. RANSAC PnP is performed on the set of 2D-3D correspondences for each detected cone, i.e. extract the keypoints by passing the patch through the keypoint regressor and use the pre-computed 3D correspondences to estimate their 3D position. One can obtain the position of each cone in the car's frame by a transformation between the camera frame and the ego-vehicle frame.

\section{Data Collection and Experiments}

\subsection{Dataset Collection and Annotation}
To train and evaluate the proposed pipeline data for object detection and keypoint regression is collected and manually-labeled. To analyze the accuracy of the position estimates using the proposed method with a single image, 3D ground-truth is collected with the help of a LiDAR.

\subsubsection{Traffic cone detection} \label{sssec:YOLO_data}
The object detector is trained on 90\% of the acquired dataset, about 2700 manually-annotated images with multiple instances of cones and performance is evaluated on 10\% of the data (about 300 unseen images).

\subsubsection{\textit{Keypoints} on cone patches}
\label{sssec:kp_data}
900 cone patches were extracted from full images and manually hand-labeled using a self-developed annotation tool.

\noindent
\textbf{Data Augmentation.} The dataset was further augmented by transforming the image with 20 random transforms. These were a composition of random rotations between [$-15^\circ$, $+15^\circ$], scaling from 0.8x to 1.5x and translation of up to 50\% of edge length. During the training procedure, the data is further augmented on the fly in the form of contrast, saturation and brightness. The final augmented annotated dataset is partitioned to have 16,000 cone patches for training and the remaining 2,000 cone patches for testing.

\subsubsection{3D ground-truth from LiDAR}
In order to test the accuracy of the 3D position estimates, corresponding object positions measured from a LiDAR are treated as ground-truth. This is done for 104 physical cones at varying distances from 4m up to 18m. The estimates are compared in Figure \ref{fig:lidar_mono} and are summarized in Section \ref{ssec:pos_acc}.

\begin{table}[!tb]
\caption{Summary of datasets collected and manually annotated to train and evaluate different sub-modules of the pipeline. Acronyms: (1) OtfA?: On-the-fly augmentation?, (2) NA: Not applicable.}
\centering
\begin{tabular}{||c | c | c | c||} 
 \hline
 Task & Training & Testing & OtfA?\\ [0.5ex] 
 \hline\hline
 Cone Detection & 2700 & 300 & Yes \\ 
 \hline
 Keypoint regression & 16,000 & 2,000 & Yes \\
 \hline
 3D LiDAR position & NA & 104 & NA \\
 \hline
\end{tabular}
\label{data_summary}
\end{table}

\begin{figure}[!tb]
    \begin{center}
        \centering
        \includegraphics[width=0.45\textwidth]{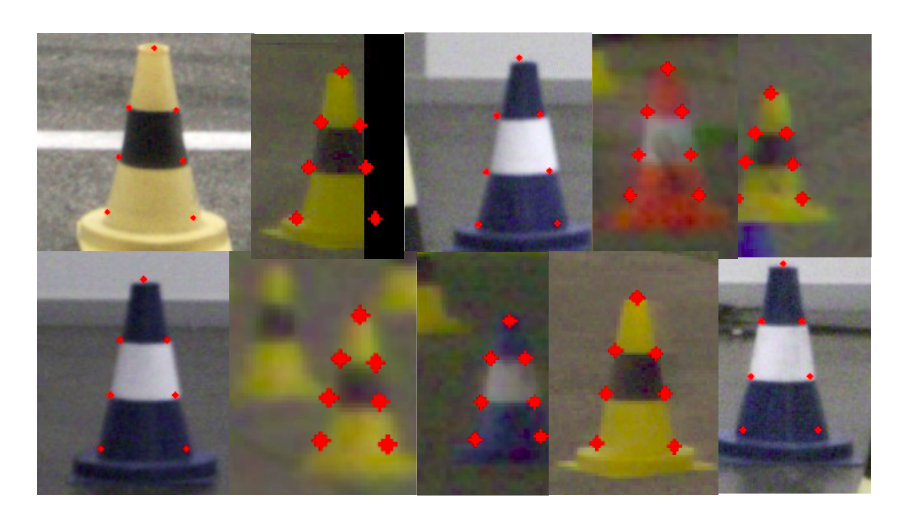}
    \end{center}
    \caption{Robust performance of keypoint regression across various scenarios. Refer to Section \ref{ssec:robust_kp_writeup} for analysis.}
    \label{fig:robust_kp}
\end{figure}

\begin{figure}[!tb]
    \centering
    \includegraphics[width=0.3\textwidth]{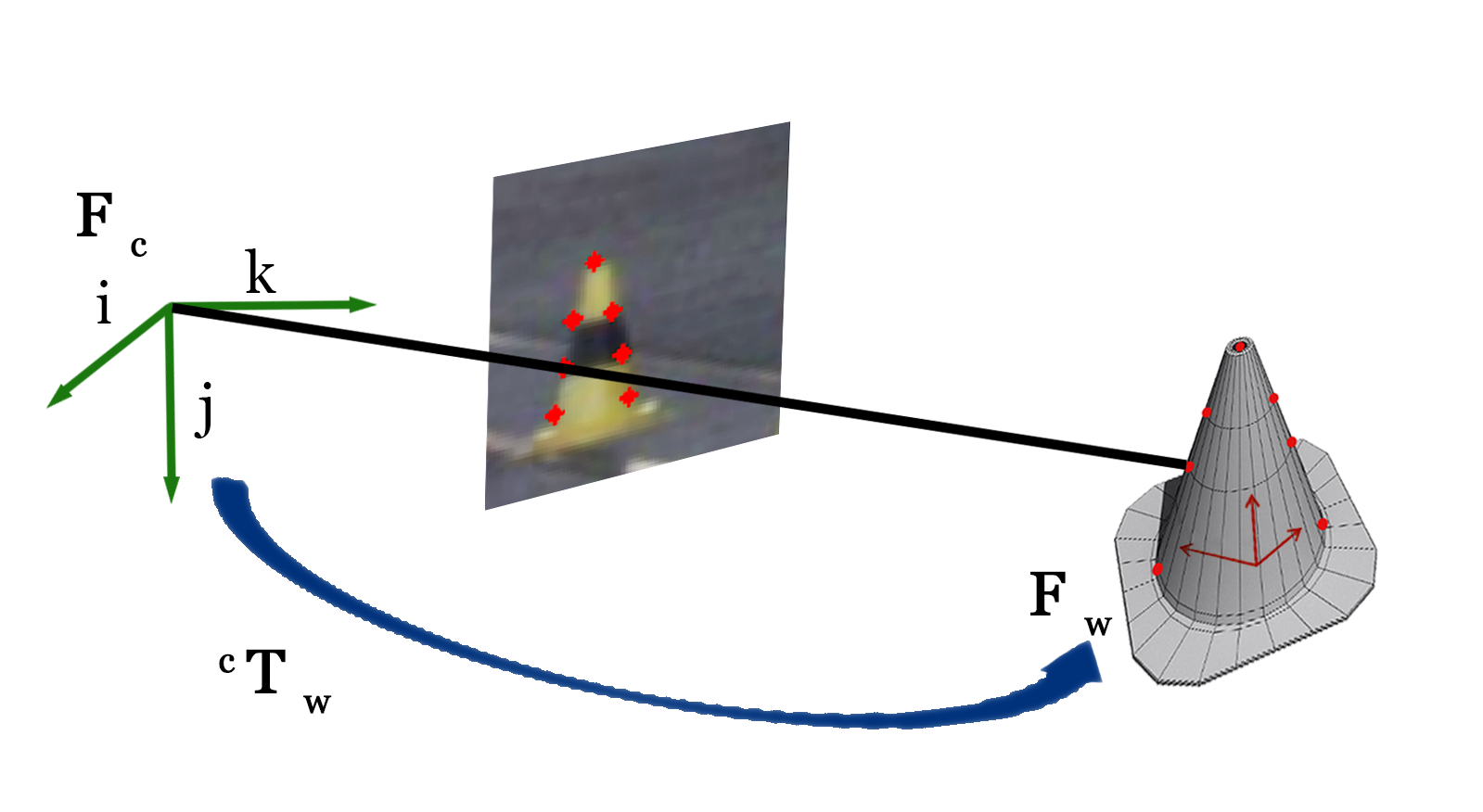}
    \caption{Schematic illustrating matching of 2D-3D correspondence and estimation of transformation between the camera frame and the world frame.}
    \label{fig:cone_pnp}
\end{figure}

This section analyzes and discusses results of the monocular perception pipeline, paying special attention to the robustness of the keypoint network and the accuracy of 3D position estimates using the proposed scheme from a single image. The keypoint network can process multiple cone patches in a single frame within 0.06s, running at 15-16 frames per second on a Jetson TX2 while running other sub-modules of the pipeline and handling 1 Gb/s of raw image data.


\subsection{Cone Detection}
Table \ref{training_testing_YOLO} summarizes the performance evaluation of the cone detection sub-module. The system has a high recall and is able to retrieve most of the expected bounding boxes. With a high precision it is averse to false detections which is of utmost importance to keep the race-car within track limits. Figure \ref{fig:robust_detection} illustrates the robustness of the cone detection pipeline in different weather and lighting conditions. The colored cone detections are shown by bounding boxes colored respectively. The key to driving an autonomous vehicle successfully is to design a perception system that has minimal or no false positives. False detections can lead to cone (obstacle) hallucination forcing the car off-course. The cone detection module is able to detect cones up to a depth of approximately 18-20m, however, it gets more consistent with cones further away due to their small size.

\begin{table}[!tb]
\caption{Performance of YOLO for colored cone object detection.}
\centering
\begin{tabular}{||c| c | c | c||} 
 \hline
  & Precision & Recall & mAP \\ [0.5ex] 
 \hline\hline
 Training & 0.85 & 0.72 & 0.78 \\ 
 \hline
 Testing & 0.84 & 0.71 & 0.76 \\
 \hline
\end{tabular}
\label{training_testing_YOLO}
\end{table}

\subsection{Keypoint Regression}
\label{ssec:robust_kp_writeup}
Figure \ref{fig:robust_kp} illustrates a montage of 10 sample patches regressed for keypoints after being detected by YOLOv2. The second cone (from the left) in the top row is detected on the right edge of the image and is only partially visible on the image and is padded with black pixels. Even with missing pixels and no information about a part of the cone, our proposed regressor predicts the keypoints. It learns the geometry and relative location of one keypoint with respect to another. Even by just partially observing a cone, it approximates where the keypoint \textit{would have been} in case of a complete image. From the examples we can see that it is able to understand spatial arrangement of the keypoints and their geometry through data. For the second cone from the left in the bottom row there is another cone in the background but the keypoint network is able to regress to the more prominent cone. One has to note that as the dimensions of the bounding box become smaller, it becomes more tricky to regress precisely due to the reduced resolution of the sub-image as can be seen in the last two cone samples in the first row here.

We train the model using the loss from equation \ref{eq:loss_fn}) which has the cross-ratio terms in addition to the mean-squared term. We evaluate the performance of our keypoint regressor using the mean-squared error. The performance on the train and test splits of the final keypoint regression model is summarized in Table \ref{training_testing_kp}. The empirical performance, measured by the mean-squared error between the prediction and ground-truth, between the train and test splits is very close meaning that the network has accurately learned how to localize keypoints given cone patches and does not overfit.

Figure \ref{fig:robust_kp} shows the robustness and accuracy of the keypoint regressor, but it represents only the internal performance of the keypoint network sub-module. In the following subsections, we analyze how outputs of intermediate sub-modules affect the 3D cone positions. We also show how variability in output of a particular sub-module ripples through the pipeline and influence the final position estimates.

\begin{table}[!tb]
\caption{Performance of the keypoint network on training and testing datasets. }
\centering
\begin{tabular}{||c|c|c||} 
 \hline
  & Training & Testing \\ [0.5ex] 
 \hline\hline
 MSE & 3.535 & 3.783 \\ 
 \hline
\end{tabular}
\label{training_testing_kp}
\end{table}

\subsection{3D position accuracy}
\label{ssec:pos_acc}
\begin{figure}[!tb]
    \centering
    \includegraphics[width=0.47\textwidth]{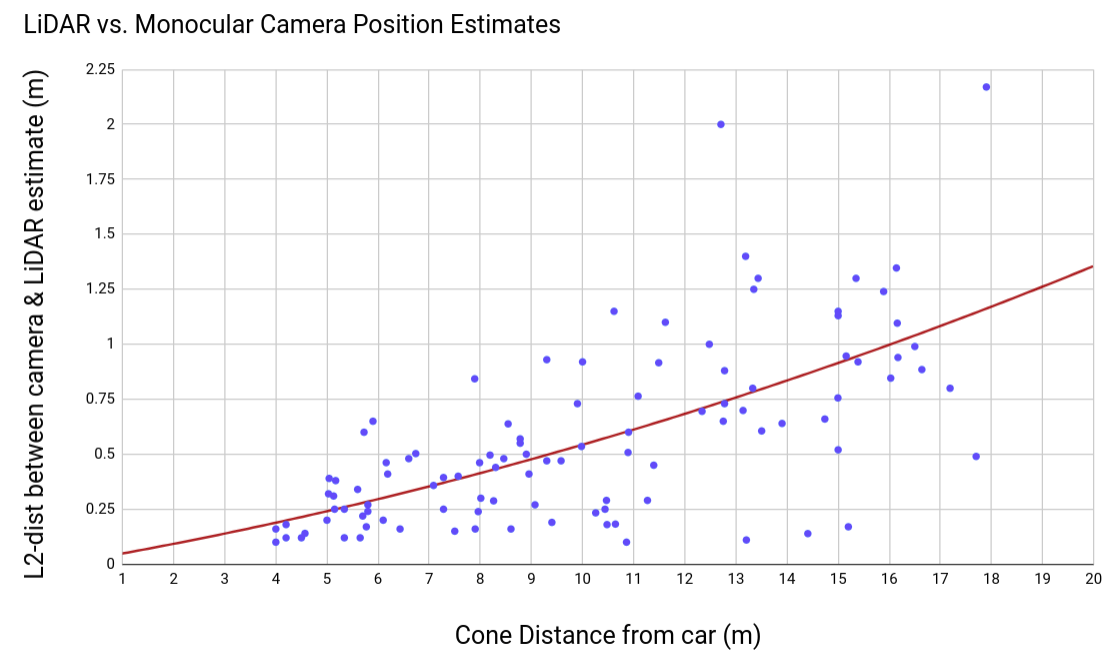}
    \caption{Euclidean distance between position estimates from LiDAR and monocular camera pipeline for the same physical cone. The x-axis represents the absolute distance of the cone from the ego-vehicle frame and the y-axis the Euclidean difference between the LiDAR and camera estimates. We fit a 2nd-degree curve (shown in red) to the points. Refer to Subsection \ref{ssec:pos_acc} for details.}
    \label{fig:lidar_mono}
\end{figure}

As it is deployed on a real-time self-driving car, one of the most crucial aspect is the accuracy of the estimated 3D positions. We compare the accuracy of the pipeline against the LiDAR's 3D estimates, which is treated as ground-truth.

Figure \ref{fig:lidar_mono} shows data from 2 different test tracks. The x-axis represents the depth, in meters, of a physical cone and along the y-axis is the Euclidean distance between the 3D position from the LiDAR estimates and the 3D position from the monocular camera pipeline. The plot consists of 104 physical cones as data points. Furthermore, a second order curve is fitted to the data, which has mostly linear components. On average, the difference is about 0.5m at a distance of 10m away from the ego-vehicle and only about 1m at a distance of 16m. At 5m the cone position is off by $\pm 5.00\%$ of its distance, and at 16m, it is off by only $\pm 6.25\%$ of its distance. The error is small enough for a self-driving car to drive on a track flanked by cones at speeds higher than 50 km/h.

\subsection{Extended perception range}
\begin{figure}[!tb]
    \centering
    \includegraphics[width=0.47\textwidth]{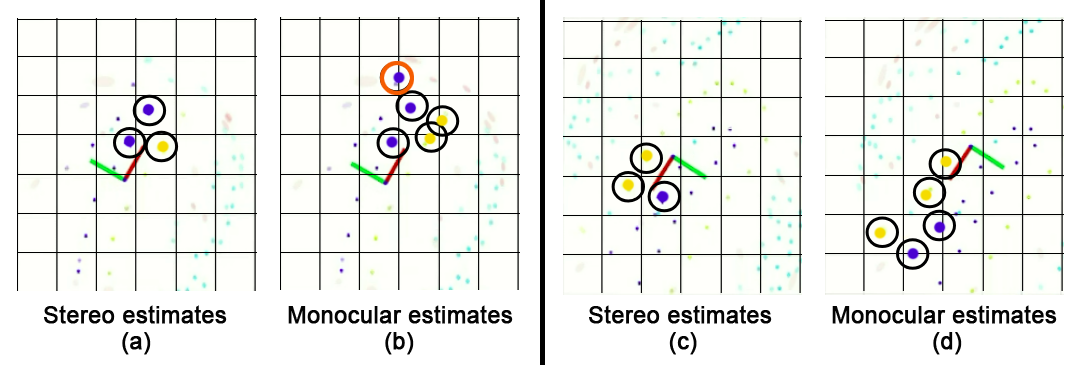}
    \caption{3D cones using computer vision are depicted as solid yellow and blue circles highlighted by black circles. In the first panel, (a) \& (b), ``gotthard driverless'', shown as a coordinate frame, with red axis pointing forward, approaches a sharp hair-pin turn. The monocular pipeline perceives a blue cone on the other side of the track (marked with an orange circle), allowing SLAM to map distant cones and tricky hair-pins. In the second panel, (c) \& (d), the car approaches a long straight. One can clearly see the difference in the range of the stereo pipeline when compared with the monocular pipeline which can perceive over an extended range of distances. With a longer perception range, the car can accelerate faster and consequently improve its lap-time. Each grid cell depicted here is $5m\times5m$.}
    \label{fig:mono_range}
    
\end{figure}

One of the goals of the method is to extend the range of perception. In Figure \ref{fig:mono_range} we compare the difference between the ranges of the monocular and the stereo pipeline. Our proposed work using a monocular camera has larger perception range than the standard triangulation solution based on stereo cameras. Additionally, the monocular camera has a longer focal length than the stereo cameras. If the stereo pair also have longer focal length, the field of view reduces introducing blind-spots where the stereo cameras cannot triangulate.  

\subsection{Effect of 2D bounding boxes on 3D estimates}

\begin{figure}[!tb]
    \centering
    \includegraphics[width=0.47\textwidth]{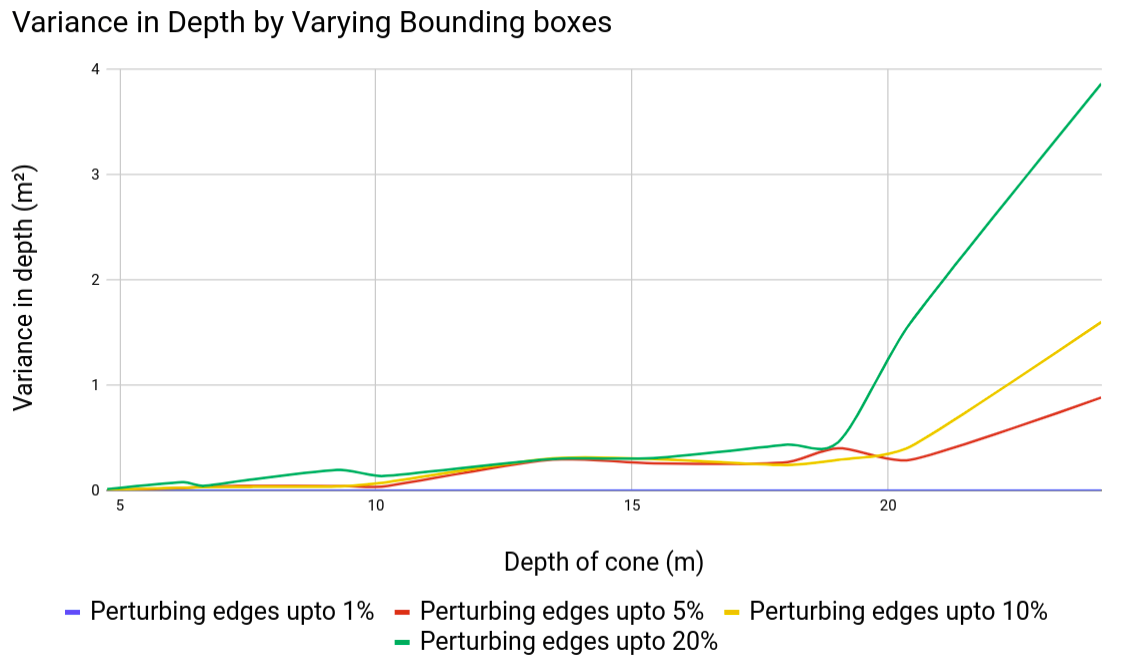}
    \caption{The variance observed in depth of cones when perturbing the dimensions and position of bounding boxes that are input to the keypoint regressor. On the x-axis is the depth of the cone while the y-axis represents the variance in the cone's depth estimate. Even with imprecise and inaccurate patches, the variance in depth estimates is quite low.}
    \label{fig:obj_det_3D_pose}
    
\end{figure}

As mentioned before, we would like to see how sub-modules have an effect on the final 3D position estimates. Here, we take a step back and analyze how variability in output from the object detection sub-module (imprecise bounding boxes) would influence the 3D positions. In this experiment, we randomly perturb the bounding box edges by an amount proportional to the height and width of the bounding box in respective directions. Due to the inherent nature of the sensor, estimating depth is most challenging using raw data from cameras. Figure \ref{fig:obj_det_3D_pose} shows how for single images, perturbing the boxes by a certain amount ($\pm 1\%$, $\pm 5\%$, $\pm 10\%$ and $\pm 20\%$) influences the variance in depth estimates.
As expected, for higher amounts of perturbation, more variance in depth estimates is observed. However, even for a $\pm 20\%$ perturbation, the variance is about 1 m$^2$ at 15m. Figure \ref{fig:obj_det_3D_pose} shows that even with imprecise and varying bounding boxes, the depth of the cone is consistent and has low variance.

For additional visualizations of the detection, regression, 3D position estimation, mapping and final navigation please visit the project page ~\url{http://people.ee.ethz.ch/~tracezuerich/TrafficCone/}

\section{Conclusion}
Accurate, real-time 3D pose estimation can be used for several application domains ranging from augmented reality to autonomous driving. We introduce a novel keypoint regression scheme to extract specific feature points by leveraging geometry in the form of the cross-ratio loss term. The approach can be extended to different objects to estimate 3D pose from a monocular camera and by exploiting object structural priors. We demonstrate the ability of the network to learn spatial arrangements of keypoints and perform robust regression even in challenging cases. To demonstrate the effectiveness and accuracy, the proposed pipeline is deployed on an autonomous race-car. The proposed network runs in real-time with 3D position deviating by only 1m at a distance of 16m. \\

\textbf{Acknowledgement.} We would like to thank AMZ Driverless for their support. The work is also supported by Toyota Motor Europe via the TRACE-Zurich project.

\bibliographystyle{ieee}
\bibliography{egbib}

\end{document}